\documentclass[lettersize,journal]{IEEEtran}
\usepackage{amsmath,amsfonts}
\usepackage{algorithmic}
\usepackage{algorithm}
\usepackage{array}
\usepackage[caption=false,font=normalsize,labelfont=sf,textfont=sf]{subfig}
\usepackage{textcomp}
\usepackage{stfloats}
\usepackage{url}
\usepackage{verbatim}
\usepackage{graphicx}
\usepackage{cite}

\usepackage{graphics} 
\usepackage{epsfig} 
\usepackage{mathptmx} 
\usepackage{amsmath} 
\usepackage{amssymb}  
\usepackage{gensymb}
\usepackage{color}
\usepackage{tabularx}
\usepackage{multirow, makecell}
\usepackage{multicol}
\usepackage{mwe,lipsum}
\usepackage{rotating}
\usepackage{tikz}
\usepackage[numbers]{natbib}
\usepackage{bbding}
\usepackage[dont-mess-around]{fnpct}
\usepackage{etoolbox}
\usepackage{subfig}

\begin{document}

\title{Depth-Adapted CNNs for RGB-D Semantic Segmentation}

\author{Zongwei Wu, Guillaume Allibert, Christophe Stolz, Chao Ma, and C\'edric Demonceaux 
\thanks{Z. Wu, C. Stolz, and C. Demonceaux are with ImViA, Université Bourgogne Franche-Comté, Dijon, France (e-mail: \{zongwei\_wu@etu.; christophe.stolz@; cedric.demonceaux@\}u-bourgogne.fr )}
\thanks{G. Allibert is with Universit\'e  C\^ote d’Azur, CNRS, I3S, Nice, France (e-mail: allibert@i3s.unice.fr)}
\thanks{Z. Wu and C. Ma are with MOE Key Lab of Artificial Intelligence, AI Institute, Shanghai Jiao Tong University, Shanghai, China (chaoma@sjtu.edu.cn)}
}

\markboth{Journal of \LaTeX\ Class Files,~Vol.~14, No.~8, August~2021}%
{Shell \MakeLowercase{\textit{et al.}}: A Sample Article Using IEEEtran.cls for IEEE Journals}


\maketitle

\begin{abstract}
Recent RGB-D semantic segmentation has motivated research interest thanks to the accessibility of complementary modalities from the input side. Existing works often adopt a two-stream architecture that processes photometric and geometric information in parallel, with few methods explicitly leveraging the contribution of depth cues to adjust the sampling position on RGB images. In this paper, we propose a novel framework to incorporate the depth information in the RGB convolutional neural network (CNN), termed Z-ACN (Depth-Adapted CNN). Specifically, our Z-ACN generates a 2D depth-adapted offset which is fully constrained by low-level features to guide the feature extraction on RGB images. With the generated offset, we introduce two intuitive and effective operations to replace basic CNN operators: depth-adapted convolution and depth-adapted average pooling. Extensive experiments on both indoor and outdoor semantic segmentation tasks demonstrate the effectiveness of our approach.
\end{abstract}

\begin{IEEEkeywords}
RGB-D Semantic segmentation, Attention, Convolution
\end{IEEEkeywords}

\section{Introduction}

As one of the fundamental tasks in computer vision, semantic segmentation aims to understand the pixel-wise label from an input image of a generic target scene. Recent advances in deep neural networks, as well as the GPU, have set new state-of-the-art (SOTA) performance in semantic segmentation. Despite significant progress in the last decade, semantic segmentation based on RGB input remains challenging in many challenging scenarios, i.e., low-contrast light, object occlusion, and separating objects sharing a similar visual appearance.

Recent developments in RGB-D sensors make RGB-D inputs accessible at a low cost, motivating research interests in designing various fusion strategies to merge multi-modal features. A number of works have demonstrated the benefit of spatial cues to improve the accuracy of semantic segmentation, affirming the effectiveness of learning from complementary modalities. In the literature, two main designs have been widely exploited: single-stream design and two-stream design. The single-stream often realizes an early fusion where RGB and depth images are simply concatenated from the input side. Different from conventional RGB networks with 3-channel, several works merge multi-modal inputs at the channel axis to form a 4-channel input (RGB-D) or 6-channel input (RGB-HHA where HHA is encoded from depth referring to disparity, height above ground, and norm angle). However, these networks directly extract features from early-mixed modalities that cannot fully explore the correlation between RGB and depth images. The two-stream strategy adopts parallel encoders that extract multi-modal features separately and further fuse them at different semantic levels. Nevertheless, compared to single-stream networks, two-stream designs inevitably increase the computational cost. Furthermore, the fusion mechanism is often pre-defined that cannot adapt to different scenarios without handcraft adjusting. 

\begin{figure}[t]
\includegraphics[width=\linewidth]{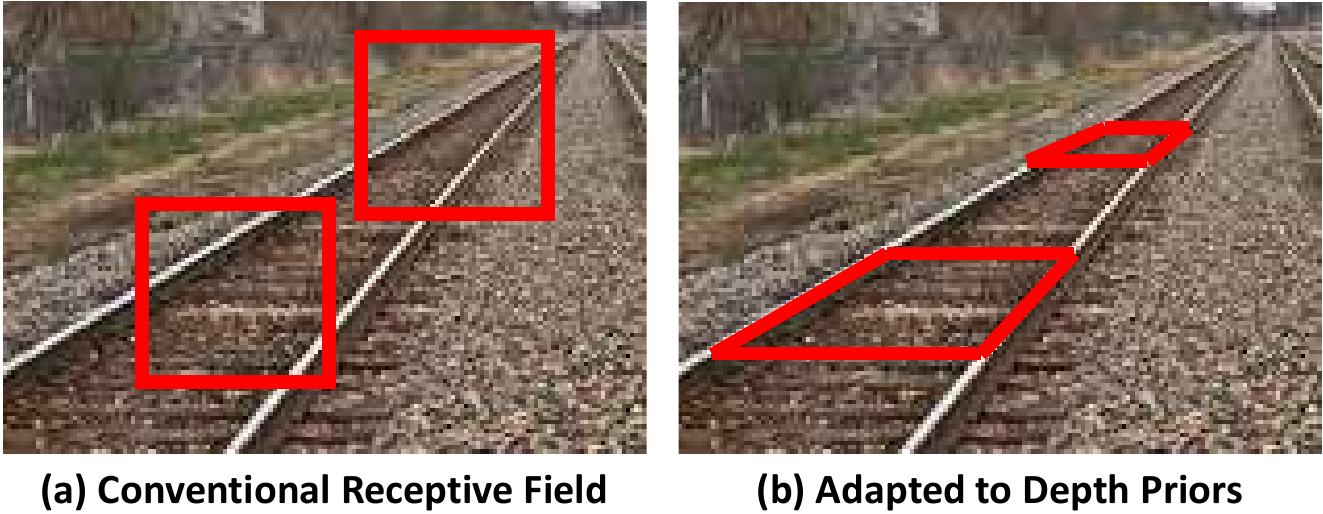}
\caption{A sketch of Depth-Adapted Sampling position. We explicitly leverage the depth priors to compute a locally-deformable sampling position, yielding a simple but efficient manner to introduce a global-local attention into CNN.}
\label{fig:adapted}
\end{figure}

In this paper, we explore differently the relationship between RGB and depth by explicitly leveraging the perspective effect. Recent non-local attention \cite{dai2017deformable,wang2018non,vaswani2017attention,woo2018cbam,hu2018senet,liu2021Swin} works in vision tasks have proved their effectiveness in modeling contextualized awareness. Several recent works \cite{chu2021conditional,pan2021integration,liu2022convnet} further figure out that aggregating both global and local attention can lead to better performance since neighboring pixels tend to have high similarity and correlation. Sharing the same idea, we seek to improve CNN with local but contextualized awareness which is fully constrained by the geometry. As shown in Figure \ref{fig:adapted}(a), the conventional convolution is designed to have a regular and fixed structure on the image plane. With additional priors on camera parameters and depth cues, we show that the sliding windows can adapt to the geometry, e.g., the vanishing effect as shown in Figure \ref{fig:adapted}(b).


Inspired by this observation, we develop a Depth-Adapted Convolutional Network, denoted Z-ACN. Z stands for the Z-axis of the camera coordinates representing the depth information. The preliminary version has been published as the conference paper \cite{wu2020depth} \footnote{Compared to the conference paper, this journal version has made the following extensions and improvements: a) We newly introduce depth-adapted average pooling (Eq. \ref{eq:avgpool}) which further improves the model performance. b) We extend our application to the pre-trained model with both VGG and ResNet encoders. c). We conduct extensive studies on both indoor and outdoor scenarios to demonstrate the effectiveness of our approach. d). We thoroughly compare with concurrent attention-based convolution to demonstrate our superior design. e). An ablation study presents the contributions of different components of our operators.}. Specifically, we propose a depth-adapted offset that can be integrated into basic functions of CNN, i.e, convolution and pooling, and introduce two new operators: depth-adapted convolution and depth-adapted average pooling.

Our proposed depth-adapted convolution replaces conventional neighboring pixels with geometrically similar ones. Concretely, we reshape the receptive field to cover pixels sharing the same 3D plane with the center of the kernel, yielding a simple but efficient manner to articulate both photometric and geometric information. The second introduced operator is depth-adapted average pooling. Sharing the same idea as the depth-adapted convolution, we re-define the notion of neighboring pixels for average pooling such that the geometrical relations will be considered while computing the mean of the local region of the feature map. Both operators break the limits on the conventional definition of neighboring pixels, forcing the network to pay attention to a larger and more malleable receptive field.

Depth-adapted operations are based on the intuition that pixels with the same geometrical character should be more likely to share the same semantic label. One common example is the vanishing effect, as illustrated in Figure \ref{fig:adapted}. We assume that pixels on the same 3D plane tend to share the same class. This 3D plane and depth variance have a high correlation. As shown in Figure \ref{fig:adapted}, we display the projection of the 3D plane of the rail on the image plane as the adapted sampling position. The depth-adapted field should be more correlated to the real scene compared to the conventional neighboring field. Essentially, our method uses depth to transform planes into a canonical pose relative to the camera, such that the extracted feature maps are also in a canonical reference frame and thus invariant to scale changes and out-of-plane rotation. The main advantages of such operations are summarized as follows:

\begin{itemize}
    \item We propose a novel depth-adapted convolutional network termed Z-ACN, that can integrate the geometric constraint into the conventional receptive field, hence improving the convolution with depth-aware contextualized attention.
    \item Our grid adaptation is processed by the non-learning method that does not introduce extra learning parameters compared to conventional counterparts.
    \item Experiments on both indoor and outdoor RGB-D semantic segmentation benchmarks demonstrate that our method can perform favorably over the baseline performance with large margins and set the new state-of-the-art performance.
\end{itemize}


\section{Related Work}

\subsection{3D Representation}

How to deal with the complementary depth is a key research topic for RGB-D semantic segmentation. Different from 2D RGB images, RGB-D images provide additional cues on 3D geometry. Therefore, a straightforward motivation is to project the 2D pixels to form the 3D representations such as voxel and point cloud. The volumetric representation \cite{3DConvLanding,wu20153d} feeds voxel data into 3D CNN. Despite the demonstrated success, it inefficiently consumes huge memory as data is often sparse on the 3D scene. Different from the voxel representation, \cite{Qi2017pointnet,NIPS2017Pointnet} propose to directly use the point cloud representation. Different 3D CNN methods are trying to adapt to the irregularity of the point cloud. \cite{li2018pointcnn} integrates an x-transformation to leverage the spatially-local correlation of point cloud \cite{Thomas2019kpconv} introduces a spatially deformable convolution based on kernel points to study the local geometry. \cite{liu2019relation} learns the mapping from geometry relations to high-level relations between points to get a shape awareness. \cite{liu2019densepoint} defines convolution as an SLP (Single-Layer Perceptron) with a nonlinear activator.

Besides, a number of efforts have been made to reduce the model complexity. \cite{tchapmi2017segcloud} adapts CRF (Conditional Random Fields) to reduce the model parameters. Multi-view methods \cite{Chen_2017MultiView,Ge20173Dpose,3DFCN,Qi2016VolMultiView} reform 3D CNN to become the combination of 2D CNNs. \cite{Chen_2017MultiView} profits from Lidar to get bird-view and front-view information in addition to a traditional RGB image. \cite{Ge20173Dpose} uses depth image to generate the 3D volumetric representation after which projections on X, Y, and Z planes are learned respectively by 2D CNNs. 3D CNN achieves better results than RGB CNN but requires further development on problems such as memory cost, data resolution, and computing time.

\subsection{2D RGB-D Fusion}

Instead of processing the 3D data, an alternative is to consider depth as another 2D image complementary to the RGB image. Deep neural networks for paired 2D RGB-D images have attracted research interests for years and numerous improvements have been achieved. At the early stage, \cite{gupta2014learning} proposes to encode a depth map to a 3-channel HHA image, which refers to Horizontal disparity, Height above ground, and normal Angle.   Afterward, the encoded HHA is widely used in RGB-D tasks since both modalities share the same dimension on both the channel axis and spatial axis \cite{hu2019acnet,Long2015FCN,xiong2020variational}. Hence, a straightforward idea is to apply identical but parallel networks on both RGB image and depth maps, and further realize data fusion at different scales  \cite{Lin2017RGBDCascaded,wang2016RGBD}. ACNet \cite{hu2019acnet} fuses RGB-D features at middle level. SA gate \cite{chen2020bi} further explicitly leverages spatial and channel cues to firstly calibrate modality in a bi-directional manner and further realize middle fusion. Despite the proven progress, the two-stream design doubles the number of encoder parameters, yielding a higher requirement on the computational cost.

To address this issue, a number of preliminary works \cite{gupta2014learning,hazirbas2016fusenet,wang2016learning} fuse the RGB-D/RGB-HHA images from the input side and form a 4- or 6-channel input with a single encoder. Based on the early fusion, D-CNN \cite{Wang2018DCNN} enhances the network with a depth similarity term which re-weight the standard convolution with the depth-related local context. Since then, various works have been developed on the forms of weight functions. \cite{chen20193d} extends the idea of \cite{Wang2018DCNN} to dilated convolution. \cite{xing20192,xing2020malleable} develop 2.5 D convolutions with a more generalized weight function. \cite{chu2018surfconv} projects 3D convolution on 2D images to form a depth-aware multi-scale 2D convolution. \cite{xiong2020variational} uses depth information to define local neighborhoods by introducing a learned Gaussian kernel. Sharing the same idea of re-weighting the convolution, ShapeConv \cite{cao2021shapeconv} integrates the channel attention into the convolution function and forms a more generalized convolution that is not limited to RGB-D context.

It can be seen that contextualized awareness has played a vital role in RGB-D fusion. For two-stream designs, multi-modal features are often firstly fed into attention module before the data fusion: \cite{li2016lstm} with ConvLSTM modules, ACNet \cite{hu2019acnet} with channel attention \cite{hu2018senet}, \cite{xiong2020variational} with a learned Gaussian convolution kernel, and \cite{chen2020bi} with a modified CBAM \cite{woo2018cbam}. For single-stream design, the contextualized awareness is directly integrated into the basic convolution function to re-calibrate the filter weight: \cite{Wang2018DCNN} with depth similarity, \cite{xing2020malleable} with a malleable depth-aware function, and \cite{cao2021shapeconv} with channel attention. Despite the popularity of attention modules in previous works, the capability of modeling long-range dependencies is still limited due to the fixed shape of the convolutional receptive field, i.e., within the 8 neighboring pixels for a conventional $3\times 3$ convolution. In contrast, we propose a depth-adapted sampling position to explicitly leverage both global and local awareness in a simple yet efficient manner. By designing a geometry-constrained offset, we aim to break the conventional receptive field to adapt to the perspective effect, yielding an effective depth-guided 2D CNN to improve the RGB understanding.

\subsection{Non-local Adaptive Model}

There are extensive surveys on attention modules in the literature. In this section, we briefly review several related deformable attention modules. In the early stages, the spatial transformer \cite{jaderberg2015spatial} aligns the feature map to alleviate the 2D rotation problem. Dilated operation \cite{yu2015Cascade} enables a larger receptive field for the convolution. Deformable-CNN (DeformCNN) \cite{dai2017deformable} learns a deformable and dense receptive fields for convolution to augment spatial sampling location. Sharing a similar idea, non-local neural network \cite{wang2018non} builds convolutional blocks with contextualized awareness.

The idea with long-range dependencies has also been proved in recent transformer works \cite{vaswani2017attention,vit,liu2021Swin} and achieves leading performance in various vision tasks. Several works \cite{chu2021conditional,liu2021Swin} discover that the global transformer attention is always a constraint around local regions, demonstrating the interest of further development in converging both global and local attention. CPVT \cite{chu2021conditional} proposes a conditional positional encoding to enhance the local awareness in the transformer backbone. Deformable Transformer \cite{deformabledetr} learns local attention with a small set of key sampling points around a reference. Swin Transformer \cite{liu2021Swin} presents a hierarchical manner to aggregate local shifted attention through different scales. Meanwhile, the recent ACmix \cite{pan2021integration} shows that self-attention and convolution are mutually beneficial and can be elegantly integrated through $1 \times 1$ convolutions. From another perspective, ConvNext \cite{liu2022convnet} re-design a ResNet with a larger receptive field and achieves better performance compared to transformer counterparts.

Despite the demonstrated promising results in previous works, we observe that the contextualized awareness are or learned through gradient descent, e.g., the positional encoding in CPVT and the offset in Deformable works, or learned through a pre-defined large receptive field, e.g., global attention in transformer and large kernel size in ConvNext. In the case of multi-modal feature learning, we seek to compute the global awareness from the additional prior. This perspective has been widely studied in the field of spherical images where the global attention is computed according to the distortion priors  \cite{cohen2018Spherical,coors2018spherenet,tateno2018distortion,CFL}. Inspired by these works, we propose to compute the non-local awareness from the depth priors, making the convolution geometry-aware for RGB-D semantic segmentation. A concurrent work SConv \cite{chen2021spatial} learns the offset from depth image. Our approach resembles the SConv in that both methods belong to depth-adapted convolution frameworks. However, one main difference is that our offset is purely defined by the geometric without requiring any gradient descent, while SConv applies convolutional layers to learn the offset from latent space. Our approach explicitly leverages the scale changes along the Z-axis of camera coordinates and out-of-the-plane rotation. Instead of adding extra learning parameters, we show that a simple and intuitive local deformation can contribute to semantic segmentation with minimal cost.


\begin{figure*}[ht]
\centering
\includegraphics[width=\linewidth,keepaspectratio]{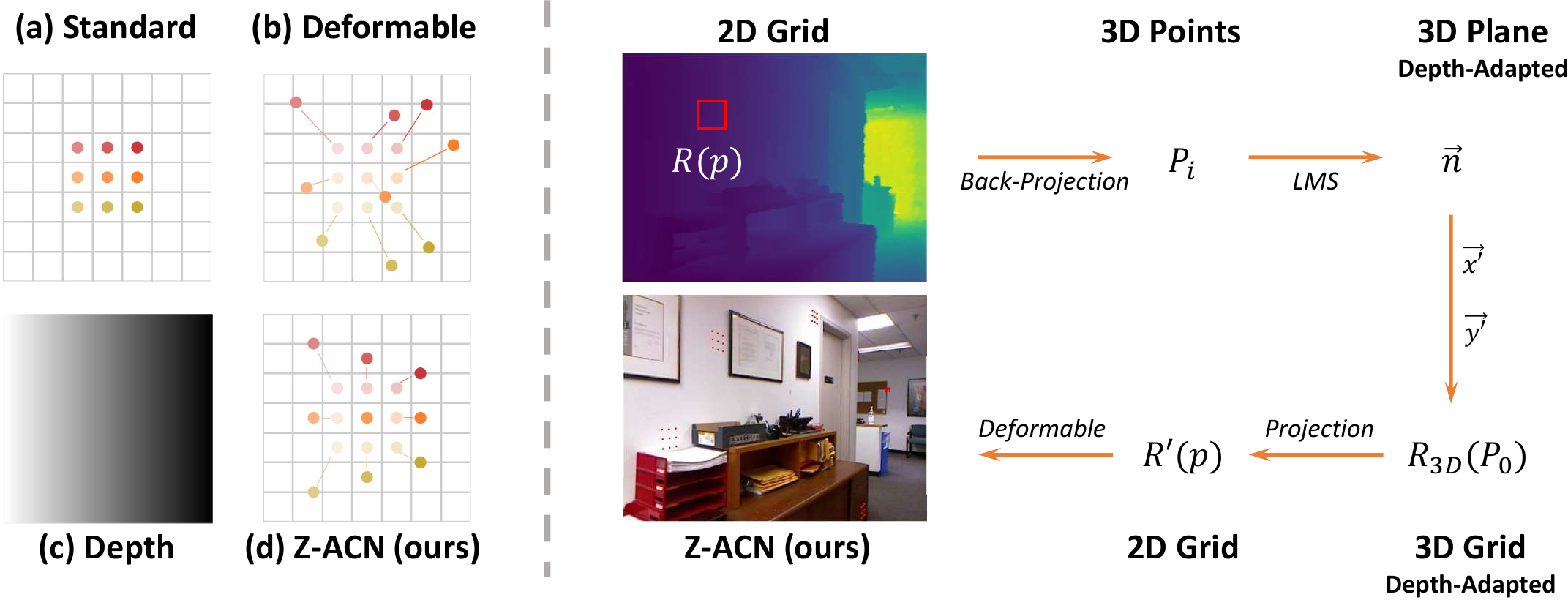}
\caption{Illustration of depth-adapted CNN. On the left we show the example of a 3 $\times$ 3 kernel: a) shows a standard 2D convolution with dilation equal to 1. b) shows the offset computed from deformable convolution \cite{dai2017deformable}. c) is the available depth data. The represented figure shows a linear change with the depth value. From left to right, the scene becomes deeper.  d) illustrates offset computed by Z-ACN which is adapted to depth. On the right, we illustrate the overview of our approach. \textbf{LMS} stands for Least Mean Square algorithm.  $(\Vec{x'}, \Vec{y'})$ are the 3D unit axis. Firstly, pixels within the 2D receptive field are back-projected to 3D space to form a point cloud, based on which a 3D plane is computed with normal $\Vec{n}$ . Secondly, a new 3 $\times$ 3 grid on the 3D space is created with the help of 3D axis $(\Vec{x'}, \Vec{y'})$ which are perpendicular to the normal $\Vec{n}$. Finally, the 3D grid is projected to the image plane, forming our depth-adapted sampling position. Zoom in for more details on the depth-guided sampling position on the RGB image.}
\label{fig:offset}
\end{figure*}

\section{Depth-Adapted Convolutional Network}

In this section, two depth-adapted operations are presented: depth-adapted convolution and depth-adapted average pooling. Figure \ref{fig:offset} shows the information propagation in our network.

First, we take a 2D conventional regular and fixed area on the depth map, which corresponds to a conventional receptive field, e.g., $3 \times 3$ convolution. We back-project the pixels to the 3D scene to get the 3D position in the camera coordinate. Second, we compute a depth-aware plane that passes through the real-world position of the kernel center and fits the best to all 3D points. Third, we create a 3D regular grid on this plane with an adapted orientation to fit the geometry. Last, we project this 3D grid on the image plane to form a 2D depth-adapted sampling grid.

Our model requires 2 inputs: input feature map and depth map (ground truth or estimated). The feature map is denoted as $\textbf{x} \in \mathbf{R}^{c_{i}\times h\times w}$, where $c_i$ is the number of input feature channel, $h$ and $w$ are the height and weight of the input feature map. The depth map is denoted as $\textbf{D} \in \mathbf{R}^{h\times w}$. $\textbf{D}$ is used to adapt the spatial sampling locations by computing the offset, denoted as $\Delta p \in \mathbf{R}^{c_{off}\times h_1\times w_1}$, where $h_1$ and $w_1$ are the height and weight of the output feature map and $c_{off} = 2\times N \times N$ for a $N\times N$ filter. Different from Deformable ConvNet, our offset does not require gradient during back-propagation. The output feature map is denoted as $\textbf{y} \in \mathbf{R}^{c_{o}\times h_1\times w_1}$, where $c_{o}$ is the number of output feature channel.

\subsection{Depth-Adapted Convolution}

A standard image convolution is formulated as: 
\begin{equation}
   \textbf{y}(p) = \sum_{p_n\in\textbf{R(p)}} \textbf{w}(p_n)  \cdot \textbf{x}(p + p_n),
\end{equation}
where \textbf{w} is the weight matrix. \textbf{R(p)} is the grid for point $p$. Physically it represents a local neighborhood on input feature map, which conventionally has regular shape with certain dilation $\Delta d$, such that : 
\begin{equation}
    \textbf{R(p)} = a \Vec{u} + b\Vec{v}
    \label{eq:2dgrid}
\end{equation}
where $(\Vec{u}, \Vec{v})$ is the pixel coordinate system of input feature map and $(a,b) \in (\Delta d \cdot \{-1, 0 , 1\})^2$.

To exploit the 3D planarity, depth-adapted convolution simply adds an adapted deformation term $\Delta p$ to adjust the spatial sampling locations : 
\begin{equation}
   \textbf{y}(p) = \sum_{p_n\in\textbf{R(p)}} \textbf{w}(p_n)  \cdot \textbf{x}(p + p_n + \Delta p_n)
   \label{eq}
\end{equation}

The convolution may be operated on the irregular positions $p_n+\Delta p_n$ as the offset $\Delta p_n$ may be fractional. To address the issue, we use the bilinear interpolation which is the same as that proposed in \cite{dai2017deformable}. In the following subsections, we will present how to process this offset from traditional computer vision algorithms.

\subsubsection{3D Planarity}

To compute the offset, firstly we assume that the camera fits the pinhole model. Therefore, with the camera parameters, we can back-project 2D pixels within the conventional field \textbf{R(p)} into camera coordinates, forming the 3D point cloud $P_i = (X_i, Y_i, Z_i)$ . An analysis of the intrinsic parameters is presented in Section \ref{study}. Let $p=(u_0, v_0)$ be the center of 2D receptive field and $P_0=(X_0, Y_0, Z_0)$ the associated back-projection on 3D space. The plane $\pi$ passing through $P_0$ and fitting the best to all $P_i$ can be extracted by applying the least square method:

\begin{equation}
\label{eq:plane}
\overrightarrow{n} = \arg\min_{\substack{    \overrightarrow{n}=(n_1,n_2,n_3)\\   ||\overrightarrow{n}||=1}} \sum_i || \overrightarrow{n} \cdot \overrightarrow{P_0 P_i}||^2
\end{equation}
where $\overrightarrow{n}=(n_1, n_2, n_3)$ is an approximation of the normal of the plane $\pi$.

Basing on the plane $\pi$, we build a new planar and regular grid, denoted as $R_{3D}(P_0)$, which is centered on $P_0$. The regular shape is defined by an orthonormal basis $(\Vec{x'}, \Vec{y'})$ on the plane $\pi$. We fix $\Vec{x'}$ as horizontal ($\Vec{x'} = (\alpha, 0, \beta)$). As $\Vec{x'}$ is on the plane $\pi$ defined by its normal $\Vec{n} = (n_1, n_2, n_3)$ and $(\Vec{x'}, \Vec{y'})$ are the orthonormal basis, we have : 
\begin{equation}
\begin{split}
    &\Vec{x'} \cdot \Vec{n} = 0; \quad ||\Vec{x'}||^2 = 1; \quad ||\Vec{n}||^2 = 1; \quad \Vec{n} \times \Vec{x'} = \Vec{y'}.
\end{split}
\end{equation}

Analytically, we can obtain $(\Vec{x'}, \Vec{y'})$ as follow:
\begin{equation}
\renewcommand\arraystretch{1.8}
\Vec{x'} = 
\begin{bmatrix}
\frac{n_3}{\sqrt{1 - n_2^2}} \\
0 \\
-\frac{n_1}{\sqrt{1 - n_2^2}}
\end{bmatrix},
\quad
\Vec{y'} = 
\begin{bmatrix}

-\frac{n_1n_2}{\sqrt{1 - n_2^2}} \\
\sqrt{1 - n_2^2}\\
-\frac{n_2n_3}{\sqrt{1 - n_2^2}}
\end{bmatrix}
\label{eq:basis}
\end{equation}
To conclude, $R_{3D}(P_0)$ is defined as : 

\begin{equation}
\label{eq:3dgrid}
R_{3D}(P_0) = a \Vec{x'} + b \Vec{y'}
\end{equation}
with $(a,b) \in (- k_u, 0 , k_u) \times (-k_v, 0 , k_v) $ where $(k_u, k_v)$ are scale factors.

A conventional 2D convolution on the image plane can be considered as realizing a planar convolution on a fronto-parallel plane on the 3D camera basis. While the depth value is constant, our depth-adapted plane $\overrightarrow{n}$ becomes the same as the fronto-parallel plane. Otherwise, our plane $\overrightarrow{n}$ can better explore the perspective effect compared to the counterpart, yielding a depth adapted sampling position $R_{3D}(P_0)$ in the camera basis.

\subsubsection{Scale Factor}

The scale factors are designed to be constant such that the 3D receptive field of each point from the feature map has the same size. In such a way, with the variance of depth, due to the perspective effect, the projected 2D receptive field on the image plane will have different sizes. The value of scale factors can be empirically set in different tasks. In our application, we want the adapted convolution performs the same as a conventional 2D convolution on a particular point $p(u_0,v_0)$ whose associated plane in Eq. \ref{eq:plane} is fronto-parallel $\{Z| Z = Z_{0}\}$. By taking into account the dilation $\Delta d$ and the camera focal length $(f_u, f_v)$, we have: 
\begin{equation}
\begin{split}
k_u = \Delta d \times \frac{Z_{0}}{f_u }  \\
k_v = \Delta d \times \frac{Z_{0}}{f_v }. 
\label{scale}
\end{split}
\end{equation}

\subsubsection{Depth-Adapted Sampling Position}

To form the depth-adapted sampling position, we denote $\textbf{R'(p)}$ as the projection of $R_{3D}(P_0)$ on the image plane : 
\begin{equation}
\begin{split}
   \textbf{y}(p) &= \sum_{p_n\in\textbf{R'(p)}} \textbf{w}(p)  \cdot \textbf{x}(p + p_n) \\
   & =  \sum_{p_n\in\textbf{R(p)}} \textbf{w}(p)  \cdot \textbf{x}(p + p_n + \Delta p_n).
\end{split}
\label{eq:zacn}
\end{equation}

Different from the conventional grid $\textbf{R(p)}$, the newly computed $\textbf{R'(p)}$ breaks the regular size and shape structure with the additional offset. In such a way, the geometry information is incorporated in RGB CNN. 

\subsection{Depth-Adapted Average Pooling}

A standard average pooling is defined as :
\begin{equation}
   \textbf{y}(p) = \frac{1}{|\textbf{R(p)}|}\sum_{p_n\in\textbf{R(p)}} \textbf{x}(p + p_n). 
   \label{eq:avgpool}
\end{equation}

This treats every pixel equally regardless of its associated geometry information, e.g. whether they belong to the same plane or not. To address this issue, similar to depth-adapted convolution, we add an extra offset to adjust the pooling field to the geometry. We force pixels sharing the same plane to contribute more to the corresponding output. For each pixel location $p_0$, the depth-adapted average pooling operation becomes :

\begin{equation}
   \textbf{y}(p) = \frac{1}{|\textbf{R(p)}|}\sum_{p_n\in\textbf{R(p)}} \textbf{x}(p + p_n + \Delta p_n). 
   \label{eq:adaptedavgpool}
\end{equation}

\subsection{Understanding Depth-Adapted operations}

In Figure \ref{fig:offset} we show several examples of depth-adapted sampling positions of given input neurons (the center) on an RGB image. We seek to profit from the depth cues to articulate both photometric and geometric information for RGB CNN. Our method integrates the geometry into the convolution by adjusting the 2D sampling grid. This pattern is integrated into Eq. \ref{eq}. In the case of conventional CNN, the shape of the grid is fixed as regular, which has difficulty adapting to the perspective effect. With the proposed Z-ACN, we can better leverage the geometric constraint in the sampling position. As shown in Figure \ref{fig:offset}, the receptive field for a closer input neuron in the 3D space is larger than that of a geometrically farther neuron. Sampling positions on the same plane also have different shapes that are adapted to the camera-projection effect. These patterns improve 2D CNN's performance with contextualized awareness without complicating the network with extra learning parameters.

\section{EXPERIMENTS}
\subsection{Experimental setup}
\label{sec:setup}

\textbf{Dataset and metrics.} We evaluate the effectiveness of our approach on both indoor and outdoor RGB-D semantic segmentation benchmarks, including NYUv2 dataset \cite{silberman2012NYUV2}, SUN RGBD dataset \cite{song2015sun} and KITTI dataset \cite{kitti}. For the NYUv2 dataset, it contains 1,449 RGB-D images which are split into 795 training images and 654 testing images. For SUN-RGBD, it contains 37 categories of objects and consists of 10,335 RGB-D images which are split into 5,285 training images and 5,050 testing images. For the KITTI dataset, we use the semantic segmentation annotation provided in \cite{xu:kitti}, which contains 70 training and 37 testing images from different scenes, with high-quality pixel annotations in 11 categories. The performance is evaluated with common metrics, i.e., Pixel Accuracy (PixelAcc), Mean Accuracy (mAcc.), Mean Region Intersection Over Union (mIoU), and Frequency Weighted Intersection Over Union (f.w.IoU).

\textbf{Implementation details.} Our approach requires paired RGB-D images as input. The depth map is first used to generate the geometry-aware offset which is further integrated into the network. As HHA encoding, the offset generation can be also realized during pre-processing since our method does not require gradient descent. We follow the same learning settings for both our proposed network and the baseline counterpart. Experiments are realized with 2 Nvidia V100 GPUs under the PyTorch framework. During inference, we apply a single-scale inference strategy.

\textbf{Comparison protocol.} We evaluate the generalization capability of our approach with different backbones, including old-fashioned VGG-16 encoders and popular ResNet encoders. We seek to demonstrate that our approach can constantly improve the baseline performance. To purely analyze the gain by applying our approach, we only replace the vanilla convolution and average pooling with our proposed depth-adapted operators.

\subsection{With VGG-16 backbone}
\begin{figure*}
\centering
\includegraphics[width=\linewidth,keepaspectratio]{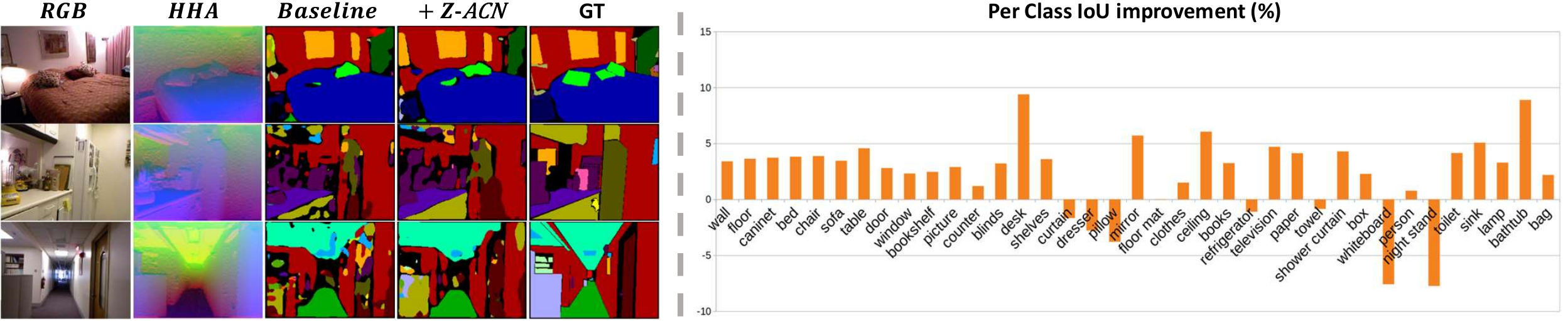}
\caption{On the left we illustrate the qualitative comparison on the NYUv2 dataset. The two first columns are the input RGB and HHA, respectively. $Baseline$ represents the semantic map obtained with early fused RGB-HHA input. + Z-ACN stands for the results obtained by inserting our depth-adapted sampling position into the baseline. It can be seen that by explicitly leveraging non-local attention, our method reasons about semantic maps closer to the ground truth (GT). The black regions in "GT" are the ignoring category. On the right we illustrate the per-class improvement above the baseline. We improve 29/37 classes with 5.2\% mean IoU increment.}
\label{fig:nyu}
\end{figure*}

\textbf{Comparison with D-CNN \cite{Wang2018DCNN}:} D-CNN is the pioneering work that integrates depth into the basic operations of convolutional networks. The depth is used to compute a similarity term to re-calibrate convolutional weight. We build our approach upon the DeepLab architecture, which is the same as D-CNN \cite{Wang2018DCNN}. Note that both the D-CNN and our approach belong to the depth-aware convolution framework. Unlike the D-CNN model, we update the depth information to break the limitation of a fixed structure, which can better leverage long-range dependencies while D-CNN seeks to refine the sampling position within the conventional receptive field. To evaluate our superior design, we follow the same training settings as D-CNN and conduct experiments on both NYUv2 and SUN-RGBD datasets. Since conventional backbones are pre-trained with RGB input, i.e., ImageNet \cite{imagenet}, which is not designed for RGB-D tasks. Hence, we follow D-CNN and train our model from scratch. We refer authors to \cite{Wang2018DCNN} for more details on the training strategies.

The quantitative comparison can be found in Table \ref{compdcnn}. We only extract features from RGB input images. The baseline model is with VGG-16 encoder under Deeplab \cite{chen2017deeplab} architecture. D-CNN stands for the performance obtained by adding depth-aware re-calibration on both convolution and pooling. Z-ACN is the result obtained with our proposed convolution and pooling where we explicitly integrate the contextualized awareness in the basic operators. Our method can achieve superior performance over the counterpart, validating the effectiveness of our depth guided sampling position which can better model geometric priors compared to D-CNN.

\begin{table}[t]
\centering
\setlength\tabcolsep{3pt}
\setlength\extrarowheight{0pt}
\caption{Comparison with D-CNN \cite{Wang2018DCNN}. Both models are trained from scratch with the same training settings. Our method achieves better performance under different datasets, show that the depth priors are better exploit with our Z-ACN.}
\begin{tabular}[ht]{ c c c c | c c c}
\hline
Dataset & \multicolumn{3}{c|}{NYUv2} & \multicolumn{3}{c}{SUN-RGBD} \\
Method  & RGB & D-CNN &  \textbf{Z-ACN} & RGB & D-CNN &  \textbf{Z-ACN} \\
\hline
PixelAcc (\%)& 50.1& 60.3 & \textbf{73.5} &66.6 & 72.4  & \textbf{78.4}  \\
mIoU (\%)& 15.9& 27.8 & \textbf{28.4} & 22.8 & 29.7  & \textbf{30.5}  \\
\hline
\end{tabular}

\label{compdcnn}
\end{table}

\textbf{Comparison with pre-trained methods}: We also evaluate our method with pre-trained weights, i.e., we initialize the weight with the pre-trained models and further fine-tune it on NYUv2 datasets. We build our approach upon the DeepLab architecture, which is the same as D-CNN \cite{Wang2018DCNN}. We report in Table \ref{pretrained} the performances of different methods. It can be seen that our method performs favorably over other methods. Compared to \cite{Qi2017Graph} which adapts a 3D CNN, our model remains a 2D CNN that requires less computational cost but achieves superior performance. Compared to our baseline, i.e., vanilla Deeplab, our Z-ACN enables significant improvements by encoding the depth information into the network. As D-CNN, our approach can also work well with the early fused RGB-HHA input, yielding a further improvement in the performance. The quantitative results validate that our operators are more effective in merging multi-modal features compared to the counterpart and set the new state-of-the-art performance with the VGG-16 encoder. 

The qualitative comparison can be found in Figure \ref{fig:nyu} which shows the improvement of our approach over the baseline. The two first columns show the input RGB image and input HHA map. $Baseline$ is the result obtained with early fused RGB-HHA input. + Z-ACN denotes that we further apply our approach over the baseline. It can be seen that our approach can favorably improve scene understanding over the counterparts by explicitly leveraging the depth cues, yielding more accurate semantic maps.

\begin{table}[t]
\centering
\caption{Quantitative comparison with VGG-16 based methods on NYUv2 dataset. Our method significantly boosts the performance over the baseline and sets a new record on VGG-16-based approaches.}

\begin{tabular}[ht]{p{2cm} | p{2.3cm} c} 
\hline

\hline
Model & Learned features & mIoU (\%)  \\
\hline
SurfConv \cite{chu2018surfconv} & RGB + HHA  & 31.0  \\ 
\hline
Eigen et al. \cite{Eigen2015MultiScale} & RGB + HHA & 34.1  \\ 
\hline
3DGNN \cite{Qi2017Graph} & RGB & 39.9  \\ 
\hline
Std2p \cite{he2017std2p} & RGB + HHA & 40.1 \\
\hline
D-CNN \cite{Wang2018DCNN}& RGB & 41.0  \\
\hline
CFN \cite{Lin2017RGBDCascaded} & RGB + HHA & 41.7  \\
\hline
D-CNN \cite{Wang2018DCNN}& RGB + HHA  & 43.9  \\
\hline
\hline
Baseline & RGB + HHA & 40.4 \\
\hline
Z-ACN (Ours)& RGB & 42.5  \\
\hline
\textbf{Z-ACN (Ours)} & \textbf{RGB + HHA } & \textbf{45.6}  \\
\hline

\hline
\end{tabular}

\label{pretrained}
\end{table}

\subsection{With ResNet backbones} 
\textbf{Plug in SOTA ESAnet \cite{esanet2021icra}}: The current SOTA CNN performance on RGB-D semantic segmentation is achieved with ESAnet. To evaluate the generalization properties of our approach, we plug our Z-ACN into ESAnet, aiming to further improve the performance with additional depth-awareness. Compared to VGG encoders, ResNet encoders are deeper with more convolutions. Hence, replacing all convolutions with depth-adapted convolutions will yield more computational cost. As suggested in previous work \cite{omniflow,shi2022panoflow}, the geometric cues play a more vital role in the first convolutional layers. Therefore, to find the best trade-off between the computational cost and the performance, we simply add a $3 \times 3$ depth-adapted convolution before the RGB encoder. This operation can be regarded as a spatial-aware positional encoding that improves the discriminability of RGB features. 

The gain by further adding our Z-ACN can be found in Table \ref{esaresnet}. With our depth-adapted operator, the new model performs favorably over the ESAnet baseline under different backbones, demonstrating the generalization capability of our approach which can easily be embedded into any existing backbones. Furthermore, since both our approach and the counterpart shares the same architecture, the improvement is purely attributed to our depth-awareness, validating the effectiveness of our geometry-guided sampling position.

\begin{table}[t]
\centering
\setlength\tabcolsep{8pt}
\setlength\extrarowheight{0pt}
\caption{Quantitative comparison with the baseline ESAnet on NYUv2 dataset. By simply adding an depth-adapted convolution, our method performs favorably over the baseline with different backbones, demonstrating the generalization capability or our Z-ACN.}
\begin{tabular}[ht]{c |l | c| c }
\hline

\hline
Backbone & Setting   & mIoU (\%)  & Improvement $\Delta$ (\%) \\
\hline
\multirow{2}{*}{ResNet-18}
& ESAnet& 46.28 & \multirow{2}{*}{0.74}\\
& Ours &\textbf{47.02} \\

\hline
\multirow{2}{*}{ResNet-34}
& ESAnet& 48.13& \multirow{2}{*}{1.02}\\
& Ours &\textbf{49.15}\\

\hline
\multirow{2}{*}{ResNet-50}
& ESAnet& 49.02&\multirow{2}{*}{1.03}\\
& Ours &\textbf{50.05}\\

\hline
\multirow{2}{*}{ResNet-101}
& ESAnet & 49.44& \multirow{2}{*}{1.76}\\
& Ours &\textbf{51.24}\\

\hline

\hline

\hline
\end{tabular}

\label{esaresnet}
\end{table}

\begin{figure*}
\centering
\includegraphics[width=\linewidth,keepaspectratio]{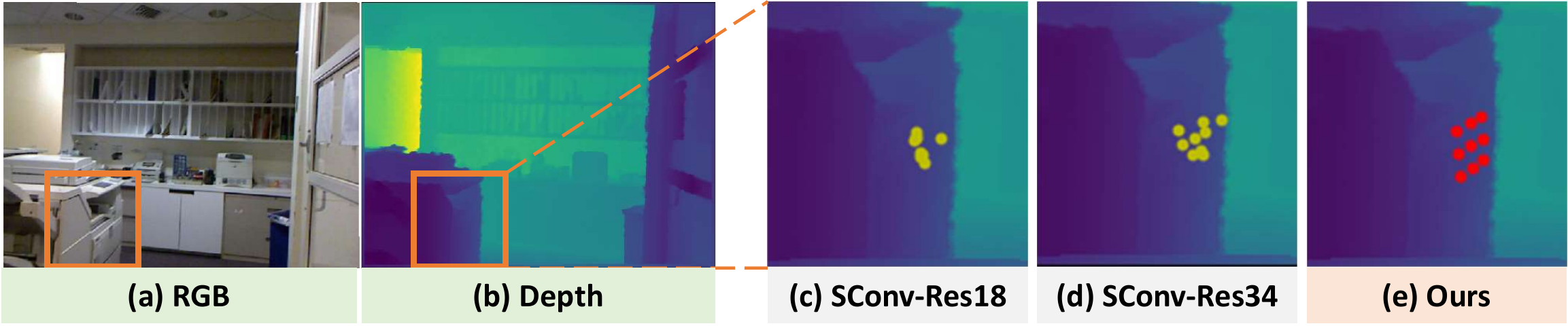}
\caption{Visual comparison with concurrent SConv \cite{chen2021spatial}. (a-b) are the input RGB and Depth. (c-d) illustrates the learned sampling position from SConv for different ResNet backbones. (e) is the receptive field computed by our approach. SConv adopts a learning diagram to generate the receptive field, resulting in different shapes for different backbones. However, our method explicitly leverages the geometric constraint for the perspective effect. Our whole process is realized without learning parameters, making the depth-adapted sampling position independent from the neural network.}
\label{fig:sconvcomp}
\end{figure*}

\textbf{Comparison with RGB-D attention convolutions}: To evaluate our Z-ACN, we compare our approach with two recent RGB-D attention convolutions, ShapeConv \cite{cao2021shapeconv} and SConv \cite{chen2021spatial}. ShapeConv decomposes the features within the receptive field into a base component and the remaining which are then calibrated with two additional learning weights before the convolution. The base component is computed by the mean function to squeeze the spatial resolution, which can be regarded as the additional channel attention for convolution. Different from ShapeConv which is not specially dedicated to RGB-D tasks, we explicitly leverage the depth prior to deform the convolutional sampling position, yielding a simple but efficient manner to integrate the spatial attention into convolution. Meanwhile, the concurrent SConv proposes a learning strategy to infer a depth-aware offset from latent space. However, for the same scene, the learned offset may vary under different settings such as different training strategies or backbones. As shown in Figure \ref{fig:sconvcomp}, while the backbone changes, SConv yields different sampling positions. Intuitively, the depth-aware offset should be only dependent on the geometry and independent of the learning factors. Different from SConv, our offset is computed without any learning parameters, making our depth-awareness constant under different environments. Further, we show through Figure \ref{fig:sconvcomp} that our computed receptive field can favorably describe the perspective effect over the counterpart. Besides, we report in Table \ref{modelsize} the model size for each method. Similar to ShapeConv, our method does not add additional parameters above the baseline and is more efficient compared to SConv which requires additional learning costs.

\begin{table}[t]
\centering
\setlength\tabcolsep{3pt}
\setlength\extrarowheight{0pt}
\caption{Model size with different attention convolutions. We choose ResNet-18 as the backbone. Similar to ShapeConv, our method do not add extra learning parameters on top of the baseline. Different from SConv, we compute the offset in a non learning manner, yielding a efficient manner to explicitly leverage the depth attention in 2D CNN.}
\begin{tabular}[ht]{c |c | c| c | c }
\hline

\hline
ResNet-18 & ESAnet \cite{esanet2021icra} & + SConv \cite{chen2021spatial} & + ShapeConv \cite{cao2021shapeconv} & + Ours  \\
\hline
Size (Mb) & 304 &+1 & +0 & +0 \\
\hline

\hline

\hline
\end{tabular}
\label{modelsize}
\end{table}

Table \ref{rescomp} illustrates the quantitative comparison with other attention convolutions. Under the consideration of a fair comparison, we embed all the operators into the ESAnet baseline and retrain them under the same settings. It can be seen that our Z-ACN outperforms the concurrent approaches with a large margin under all backbones. This highlights the effectiveness of our depth-constraint attention compared to channel attention (ShapeConv) and learned depth attention (SConv).

\begin{table}[t]
\centering
\setlength\tabcolsep{4pt}
\setlength\extrarowheight{0pt}
\caption{Quantitative comparison with other attention convolution methods on NYUv2 dataset. All methods are implemented on the ESAnet baseline and trained under the same settings. Our approach achieves better mIoU compared to concurrent works under different backbones, validating the effective of our geometry-constrained sampling position.}
\begin{tabular}[ht]{c |l | c| c | c |c }
\hline

\hline
Backbone & Setting & PixelAcc & mAcc & mIoU & f.w.IoU  \\
\hline
\multirow{3}{*}{ResNet-18}

& +SConv &74.19 & 60.01 & 46.93 & 60.26 \\
& +ShapeConv & 74.11 & 59.37 & 46.38 & 60.61\\
& +Ours & \textbf{74.35} & \textbf{59.82} & \textbf{47.02} & \textbf{60.73}\\
\hline
\multirow{3}{*}{ResNet-34} 

& +SConv & 74.95 & 61.08 & 47.99 & 61.77\\
& +ShapeConv & 74.68 & 61.07 & 47.70 & 61.20\\
& +Ours & \textbf{75.78} & \textbf{62.81} & \textbf{49.15}& \textbf{62.64}\\
\hline
\multirow{3}{*}{ResNet-50}

& +SConv & 76.13 & 62.36 & 49.04 & 63.00 \\
& +ShapeConv & \textbf{76.17} & 62.45 & 49.58 & \textbf{63.02}\\
& +Ours & 75.88 & \textbf{63.55} & \textbf{50.05} & 62.99\\
\hline
\multirow{3}{*}{ResNet-101} 

& +SConv & 76.49 & 63.65 & 50.43 & 63.67 \\
& +ShapeConv & 76.45 & 63.28 & 50.10 & 63.46\\
& +Ours &\textbf{77.00} & \textbf{64.26} & \textbf{51.24} & \textbf{64.32}\\

\hline

\hline

\hline
\end{tabular}

\label{rescomp}
\end{table}

\textbf{Comparison with SOTA performance}:
We compare the performance of our Z-ACN with other state-of-the-art models. The quantitative results can be found in Table \ref{tab:nyu}. Our Z-ACN sets the new state-of-the-art performance in NYUv2 datasets. Compared to ShapeConv when both methods use ResNet-101 as the backbone and adopt single-scale inference, our approach achieves 3.8\% mIoU improvement. 


\begin{table}[t]
\caption{Performance comparison with SOTA methods on NYUv2 dataset. $\star$ denotes the multi-scale strategy. Our method is tested with single-scale inference strategy and sets the new state-of-the-art performance among ResNet based models.}
\label{tab:nyu}
\centering
\setlength\tabcolsep{1pt}
\setlength\extrarowheight{0pt}
\begin{tabular}[ht]{l c c c c c }
\hline

\hline

\hline

\hline
 Method  & Backbone & PixelAcc & mAcc & mIoU & f.w.IoU \\
\hline
$ACNet$ \cite{hu2019acnet}  &ResNet-50 & - & - & 48.3 & - \\
$2.5D$ \cite{xing20192}  &ResNet-101& 75.9 & - & 49.1 & - \\
$ShapeConv$ \cite{cao2021shapeconv} &ResNet-101 & 74.5 & 59.5 & 47.4 & 60.8 \\
\hline

\hline

\hline

$\star CFN$ \cite{Lin2017RGBDCascaded}  &ResNet-152 & - & - & 47.7 & - \\
$\star 3DGNN$ \cite{qi20173d} &ResNet-101 & - & 55.7 & 43.1 & - \\
$\star RDFNet$ \cite{park2017rdfnet} &ResNet-152 & 76.0 & 62.8 & 50.1 & - \\
$\star ShapeConv$ \cite{cao2021shapeconv} &ResNet-101 & 75.5 & 60.7 & 49.0 & 61.7 \\
$\star Malleable$ \cite{xing2020malleable}  &ResNet-101 & 76.9 & - & 50.9 & - \\
$\star SGNet$ \cite{chen2021spatial}  &ResNet-101 & 76.8 & 63.1 & 51.1 & - \\
$\star CANet$ \cite{zhou2020canet} &ResNet-101 & 76.6 & 63.8 & 51.2 & - \\
\hline

\hline

\hline
Z-ACN (Ours)  &ResNet-101 & \textbf{77.0} & \textbf{64.3} & \textbf{51.2} & \textbf{64.3} \\
\hline

\hline

\hline

\hline
\end{tabular}
\end{table}

\textbf{Outdoor scene}:
\label{outdoor}
We also evaluate our approach to the outdoor scene, e.g., KITTI \cite{kitti}. The vanilla KITTI dataset provides RGB and lidar input. We take the dataset presented in \cite{xu:kitti} which provides a dense depth map. We validate all methods on the held-out testing set due to the smaller size and lack of a proposed validation split.

We adopt the same modified ResNet-18 as presented in \cite{chu2018surfconv} as our backbone with skip-connected fully convolutional architecture \cite{Long2015FCN}. The conventional convolution is replaced by our proposed operator. 

Our model is compared with 3D representation such as PointNet \cite{Qi2017pointnet}, Conv3D \cite{tchapmi2017segcloud,song2017semantic} and 2D representation such as DeformCNN \cite{dai2017deformable} and SurfConv \cite{chu2018surfconv}. Conv3D \cite{tchapmi2017segcloud,song2017semantic} and PointNet \cite{Qi2017pointnet} use the hole-filled dense depth map provided by the dataset to create 3D input. For PointNet, the source code is used to use RGB plus gravity-aligned point cloud (pcl). The recommended configuration \cite{Qi2017pointnet} is used to randomly sample points. The sample number is set to be 25k. For Conv3D, the SSCNet architecture \cite{tchapmi2017segcloud} is used and is trained with flipped - TSDF and RGB. The resolution is reduced to $240\times144\times240$ voxel grid. For DeformCNN, RGB images and HHA images are chosen as input for a fair comparison.  For SurfConv, we compare with their best performance, which requires a resampling on the input image to be adapted to the 8 levels of depth. For all the above-mentioned models, we follow the same configuration and learning settings as discussed in \cite{chu2018surfconv}.

The quantitative result is reported in Table \ref{scratchkitti} that all methods are trained from scratch following \cite{chu2018surfconv}. While dealing with an outdoor scene, 3D methods such as point cloud suffer from computational costs compared to 2D CNN which extracts features from images. It is also the case for Conv3D \cite{tchapmi2017segcloud,song2017semantic} since voxelizing the whole 3D space is time-consuming. Compared to these 3D methods, our model remains 2D CNN but achieves a better result. DeformCNN \cite{dai2017deformable} takes into RGB + HHA as input and learns offsets to deform the sampling position. Nevertheless, the offset is learned from the input feature maps which do not explicitly leverage the geometric constraints. In contrast, our model computes the offset from low-level constraint, i.e., 1-channel depth, with traditional algorithms and does not require gradient descent. The result in Table \ref{scratchkitti} shows that our model performs favorably over DeformCNN without extra learning parameters, validating the effectiveness of our depth-adapted sampling position. SurfConv is a concurrent work that incorporation 3D information into 2D CNN. However, it requires additional pre-processing on the input data such that depth-guided image resampling. Instead, we encode the depth into the CNN via the bias of offset. Compared to the concurrent method, our approach achieves large performance gains.

\begin{table}[t]
\centering
\caption{Comparison on KITTI test set. Our methods achieve better performance compared to 3D approaches and the concurrent SurfConv. It is worth noting that with single RGB input, our depth-adapted sampling position enables significant improvement over our baseline, validating the effectiveness of depth-guided non-local attention. Models are trained from scratch.}
\begin{tabular}[ht]{p{2cm}| p{2.2cm} c c  }
\hline

\hline
Model & Learned features & Acc (\%)  & mIoU (\%) \\
\hline
PointNet \cite{Qi2017pointnet} &  RGB + pcl &55.1 & 9.4\\
Conv3D \cite{tchapmi2017segcloud,song2017semantic} & RGB + voxel &  64.5 & 17.5\\ 
DeformCNN \cite{dai2017deformable} &  RGB + HHA & 79.2& 34.2 \\
SurfConv-8 \cite{chu2018surfconv} & RGB + HHA & 79.4& 35.1 \\
\hline
\hline
Baseline & RGB &  79.3   & 31.3  \\
Z-ACN (Ours)  & RGB&  79.7   & 33.5  \\
Z-ACN (Ours)& \textbf{RGB + HHA} &  \textbf{80.1}  & \textbf{35.8} \\
\hline

\hline
\end{tabular}

\label{scratchkitti}
\end{table}

\begin{figure*}
\centering
\includegraphics[width=\linewidth,keepaspectratio]{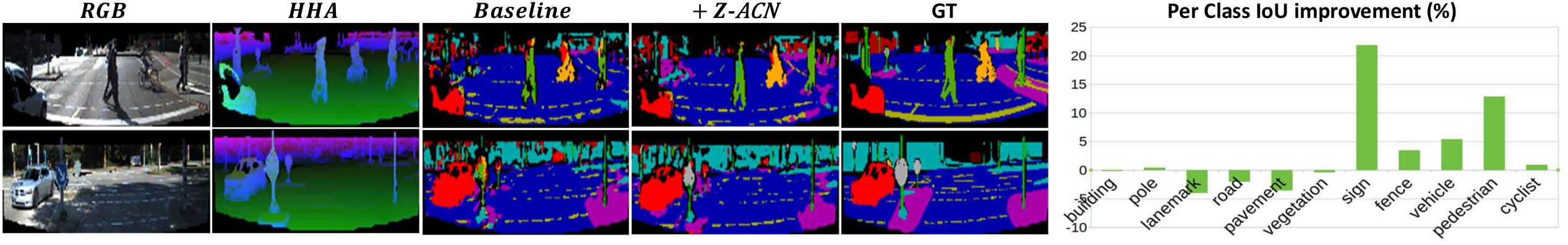} 
\caption{On the left we illustrate the qualitative comparison on the NYUv2 dataset. The two first columns are the input RGB and HHA, respectively. $Baseline$ represents the semantic map obtained with early fused RGB-HHA input. + Z-ACN stands for the results obtained by inserting our depth-adapted sampling position into the baseline. It can be seen that our approach can also improve the baseline performance in outdoor scenes. The black regions in "GT" are the ignoring category. On the right, we illustrate the per-class improvement above the baseline. We improve 7/11 objects with 3.7\% mean IoU increment. Segmentation results on the KITTI test dataset. GT stands for ground truth. The black regions in "GT" are the ignoring category.}
\label{fig:kitti}
\end{figure*}

We present in Table \ref{prekitti} the quantitative comparison over the baseline with weight initialization. $Baseline_1$ and $Baseline_2$ represent the result obtained with RGB input and early fused RGB-HHA input, respectively. + Z-ACN stands for the results obtained by inserting our depth-adapted offset into the baseline. It can be seen that our methods can significantly enable gains over the baseline performance with improved depth-awareness. We illustrate in Figure \ref{fig:kitti} the per-class IoU improvement with RGB input. Compared to the baseline, our approach enables improvement on 7/11 objects, especially "salient" objects in the urban scene such as the vehicle, cyclists, and pedestrians. However, we also observe that our approach achieves lower performance in detecting lanemark. This is because the lanemark is co-planar as the road that the confusing geometrical information may add noises for our depth-adapted model.

\begin{table}[t]
\centering
\caption{Quantitative comparison on KITTI test set. Networks are trained from pre-trained models.}
\begin{tabular}[ht]{ c c c | c c}
\hline
KITTI & $Baseline_1$ &  + Z-ACN & $Baseline_2$   & \textbf{+ Z-ACN} \\
\hline
mAcc (\%)&  48.3 & \textbf{49.5} & 51.8   & \textbf{55.1}\\
mIoU (\%)&  39.1 & \textbf{40.6} & 41.6   & \textbf{45.3} \\
\hline
\end{tabular}

\label{prekitti}
\end{table}

The qualitative comparison over the baseline is shown in Figure \ref{fig:kitti}. The two first columns show the input RGB image and input HHA map. $Baseline$ is the result obtained with early fused RGB-HHA input. + Z-ACN denotes that we further replace the baseline convolution with our approach. By explicitly leveraging the geometry, our approach constrains the network to pay more attention to boundaries and reason about semantic maps with higher accuracy. We observe that objects like the vehicle, pedestrian, and cyclist are better segmented, as well as the sign. Recognizably, these objects do not share the same depth compared to the background (road or pavement). Hence, our adapted sampling position contributes to improving the discriminability of these salient objects.

\section{Additional Studies and Discussions}
\label{study}
In this section, we conduct additional studies on the NYUv2 dataset to validate the efficiency, robustness, and flexibility of our operators. We choose VGG-16 with Deeplab as the baseline. The features are extracted from RGB images. The depth map is used to guide the sampling position. 

\subsection{Intrinsic Parameters}

Our model requires the intrinsic parameters to back-project the pixels to the 3D scene and project the depth-adapted 3D planar grid to the image plane. This pattern is integrated into Eq. \ref{eq:plane} and Eq. \ref{scale}. Demanding camera parameters as priors can be a strong assumption that limits the application. Therefore, we evaluate the performance with a randomly set camera matrix.

The intrinsic parameters include the principal point and the focal length. However, most models resize the input image shape, which results in difficulties in using the official principal point value. Hence, we assume that the principal point is the same as the center of the input image and chose a random value for focal length. We set $(f_u, f_v) = (100, 100)$ for NYUv2 dataset, where the official value is around $(519, 519)$. We retrain the new model under the same training setting. The quantitative result is reported in Table \ref{intrinsic}. We denote $k_r$, the result obtained with randomly chosen intrinsic parameters, and $GT$, the result obtained with official values.

\begin{table}[t]
\centering
\setlength\tabcolsep{10pt}
\setlength\extrarowheight{0pt}
\caption{Empirical analysis on the influence of the intrinsic parameters. All methods are trained from pre-trained model under the same setting.}
\begin{tabular}[ht]{ l c c }
\hline

\hline
NYUv2 (\%)    & mAcc   &  mIoU    \\
\hline
Baseline &  51.9 &  40.4  \\
Z-ACN ($k_r$) &53.4 &  41.6\\
Z-ACN (GT) & \textbf{55.2} &  \textbf{42.5}  \\
\hline

\hline
\end{tabular}

\label{intrinsic}
\end{table}

It can be seen that with an arbitrary value for intrinsic parameters, our model can still achieve favorable performance compared to the baseline. Compared to the result obtained with GT intrinsic value, the loss is only 0.2\% for mAcc and 0.9\% for mIoU. The result validates that our model can get rid of the assumption of the input intrinsic parameters under the condition that they are logically chosen.

\subsection{Ablation Study} 

To further verify the functionality of both depth-adapted convolution and depth-adapted average pooling, the following experiments are conducted. 
\begin{itemize}
    \item For results trained from scratch: we analyze a) baseline performance, b) a deep layer convolution replaced by Z-ACN, c) first convolutions from all layers replaced by Z-ACN, d) CNN replaced by Z-ACN including the average pooling.
    \item For results trained from pre-trained weight: we analyze a) baseline performance, b) the first convolution from a deep layer replaced by Z-ACN, c) the second convolution from a deep layer replaced by Z-ACN, d) the third convolution from a deep layer replaced by Z-ACN, e) CNN replaced by Z-ACN including the average pooling.
\end{itemize}

\begin{table}[t]
\centering
\setlength\tabcolsep{10pt}
\setlength\extrarowheight{0pt}
\caption{Results of using depth-adapted operators in different layers. Experiments are conducted on NYUv2 test set. $i$ stands for the number of convolution layers.}
\begin{tabular}[ht]{ l |l |c } 
\hline

\hline
 & Configuration & mIoU (\%)  \\
\hline
\multicolumn{3}{c}{Result from scratch} \\
\hline
a) & Baseline & 24.0  \\ 
\hline
b) & Z-Conv5\_1 & 27.6  \\ 
\hline
c) & Z-Convi\_1 & 29.7  \\ 
\hline
d) & \textbf{Z-Convi\_1 + Z-AvgPool} & \textbf{30.4}  \\ 
\hline
\multicolumn{3}{c}{Result from pre-trained} \\
\hline
a) & Baseline & 40.4  \\
\hline
b) & Z-Conv5\_1 & 42.2 \\
\hline
c) & Z-Conv5\_2 & 41.7 \\
\hline
d) & Z-Conv5\_3 & 41.7 \\
\hline
e) & \textbf{Z-Conv5\_1 + Z-AvgPool} & \textbf{42.5} \\
\hline

\hline
\end{tabular}

\label{ablation}
\end{table}

Experimental results are reported in Table \ref{ablation}. While learning from scratch, our operators can effectively extract features with geometric relationships and improve the segmentation performance. By comparing (a) and (b), we only replace deep convolution with our approach, i.e., the first convolution of layer 5 of VGG-16, we achieve a 3.6\% gain on mIoU.  (c) illustrates the result with the first convolution of all layers replaced by our approach.  Our Z-ACN enables a 5.7\% gain compared to the baseline (a). Finally, by introducing the depth-adapted average pooling (d), we observe that the performance can be further promoted, validating the effectiveness of our depth-adapted pooling method.

While learning from the pre-trained model, we firstly want to argue that the existing weight may not be fair nor suitable for our adapted convolution. The existing weight is learned with a fixed size and shape structure, while our adapted convolution breaks this limitation. The most suitable pre-trained weight for our operator might require training our depth-adapted model on ImageNet, which is impossible since the depth information is not available on this dataset.

Nevertheless, we still show that our approach can benefit from the conventional pre-trained weights. By fine-tuning the weights, Table \ref{ablation} illustrates that replacing the first convolution from a deep layer contributes the most to the performance by 1.8\% over the baseline. By introducing the depth-adapted average pooling, the performance can be further promoted.

\section{CONCLUSIONS}

In this paper, we propose a novel 2D CNN to include geometric information in RGB CNN. Different from previous works that integrate the channel or spatial attention into convolution through learning methods, our network fully leverages the geometric constraint additional training parameters, making the depth-awareness independent of the learning settings. We introduce two basic depth-adapted operators that can be easily integrated into the existing CNN model. Extensive studies demonstrate the generalization properties of our methods which perform favorably over the baseline and other convolutions. Experiments on challenging RGB-D datasets demonstrate that our approach performs well over the state-of-the-art methods by large margins. In future works, we will try to extend the application to other popular tasks instance segmentation, object detection, or even passing from image to 3D data.

\section*{ACKNOWLEDGMENT}
This work was supported by the French National Research Agency through ANR CLARA (ANR-18-CE33-0004) and financed by the French Conseil R\'egional de Bourgogne - Franche - Comt\'e.

\bibliographystyle{IEEEtran}
\bibliography{egbib}

\vfill

\end{document}